\documentclass[final]{article}

\usepackage[utf8]{inputenc}
\usepackage[T1]{fontenc}
\usepackage{hyperref}
\usepackage{url}
\usepackage{booktabs}
\usepackage{amsfonts}
\usepackage{amsmath}
\usepackage{amssymb}
\usepackage{amsthm}
\usepackage{nicefrac}
\usepackage{microtype}
\usepackage{graphicx}
\usepackage{natbib}
\usepackage{algorithm}
\usepackage{algorithmic}
\usepackage{xcolor}
\usepackage{subcaption}
\usepackage{multirow}

\newtheorem{conjecture}{Conjecture}
\newtheorem{proposition}{Proposition}

\title{Learning to Perturb Hidden Representations for Generalizable Deep Learning}

\author{
  Hua Li\\
  Henan University\\ 
  \texttt{lihuahenu@163.com} 
}

\begin{document}

\maketitle

\begin{abstract}
Deep neural networks process data through a cascade of representations: input features, hidden activations, logits, and loss. While perturbations at the input, logit, and label levels have been systematically studied, the intermediate hidden activations, which constitute the bulk of the network's computation, have received no unified perturbation analysis. In this paper, we establish a unified framework for hidden activation perturbation, revealing that Dropout, Manifold Mixup, adversarial feature perturbation, and related methods all impose specific forms of activation perturbation but with class-agnostic or random strategies. We conjecture that expansive perturbation (increasing activation norm) acts as positive augmentation, while contractive perturbation (decreasing activation norm) acts as negative augmentation, and that the perturbation layer determines whether the effect resembles input-level augmentation (shallow layers) or logit-level manipulation (deep layers). We propose Learning to Perturb Activations (LPA), which adaptively perturbs activations at a selected hidden layer with class-level perturbations learned via PGD. We further provide theoretical analysis connecting activation perturbation to flat minima and perturbation amplification through layers. Experiments on balanced classification, long-tail classification, and domain generalization demonstrate that LPA consistently outperforms existing methods and provides complementary benefits to logit perturbation methods such as LPL.
\end{abstract}

\section{Introduction}
\label{sec:intro}

In the forward pass of a deep neural network, data flows through a sequence of transformations: input features, hidden activations across multiple layers, logits, and finally the loss. Perturbation at various points along this pipeline has been shown to improve generalization, robustness, and calibration. At the input level, feature perturbation methods such as data augmentation~\citep{zhang2018mixup} and adversarial training~\citep{madry2018towards} modify the input or its representations. At the logit level, methods including Label Smoothing~\citep{szegedy2016rethinking}, Logit Adjustment~\citep{menon2021longtail}, and Logit Perturbation Learning (LPL)~\citep{wei2022logit,delving2024li,li2023class} perturb the pre-softmax outputs. At the label level, Mixup~\citep{zhang2018mixup} and its variants interpolate between training labels.

However, the hidden activations that occupy the vast majority of network layers have not been examined through a systematic perturbation lens. This gap is notable because several widely used methods implicitly perturb hidden activations. Dropout~\citep{srivastava2014dropout} zeros out random activation units, DropBlock~\citep{ghiasi2018dropblock} applies structured zeroing, Shake-Shake~\citep{gastaldi2017shake} and ShakeDrop~\citep{yamada2019shakedrop} randomly scale residual branch activations, Manifold Mixup~\citep{verma2019manifold} interpolates activations between samples, and adversarial feature perturbation~\citep{volpi2018adversarial} perturbs features to maximize the loss. These methods are motivated by different objectives (regularization, robustness, generalization) and have been developed independently without a unifying framework.

Analogous to LPL~\citep{wei2022logit}, which establishes the correspondence between logit perturbation and positive/negative data augmentation, we observe a similar correspondence for hidden activation perturbation: expansive perturbation (increasing activation norm) effectively generates more diverse representations for a class, acting as positive augmentation, while contractive perturbation (decreasing activation norm) compresses the class's representation distribution, acting as negative augmentation. Moreover, the choice of perturbation layer determines the nature of the effect: shallow-layer perturbation influences low-level feature diversity (similar to input augmentation), while deep-layer perturbation affects semantic class relationships (similar to logit perturbation).

A key distinction from logit perturbation is the dimensionality of the perturbation space. Logit perturbation operates in $\mathbb{R}^C$ where $C$ is the number of classes, providing coarse category-level control. Activation perturbation operates in $\mathbb{R}^{d_l}$ where $d_l$ is the hidden layer dimension, typically much larger than $C$. This higher-dimensional space enables finer control over intra-class representation structure that logit perturbation cannot achieve.

Based on these observations, we propose Learning to Perturb Activations (LPA), which adaptively perturbs activations at a selected hidden layer with class-level perturbations. The perturbation direction and magnitude are learned via Projected Gradient Descent (PGD), guided by whether each class requires positive or negative augmentation.

Our main contributions are:
\begin{enumerate}
    \item We establish a unified framework for hidden activation perturbation, showing that Dropout, Manifold Mixup, adversarial feature perturbation, and related methods can all be understood as specific instantiations with class-agnostic strategies.
    \item We propose LPA, which adaptively perturbs hidden-layer activations at the class level with layer-adaptive perturbation bounds, and provide theoretical analysis connecting activation perturbation to flat minima and perturbation amplification.
    \item We analyze the layer-dependence of perturbation effects and propose a layer selection strategy, showing that LPA generalizes LPL as a special case when the perturbation layer is the logit layer.
    \item We conduct experiments on three scenarios (balanced classification, long-tail classification, and domain generalization), demonstrating that LPA consistently improves upon existing methods and provides complementary benefits to LPL.
\end{enumerate}

\section{Related Work}
\label{sec:related}

\subsection{Regularization via Activation Modification}

Dropout~\citep{srivastava2014dropout} and its variants zero out random activation units during training, serving as an implicit form of activation perturbation. Spatial Dropout~\citep{tompson2015spatial} drops entire feature maps, and DropBlock~\citep{ghiasi2018dropblock} extends this to contiguous regions. Stochastic Depth~\citep{huang2016stochastic} randomly drops entire residual blocks, and R-Drop~\citep{liang2021rdrop} enforces consistency between two Dropout passes. Shake-Shake~\citep{gastaldi2017shake} and ShakeDrop~\citep{yamada2019shakedrop} randomly scale residual branch activations. These methods all perturb activations, but the perturbation form and location are fixed and not adaptive to class-level characteristics.

\subsection{Feature-level Data Augmentation}

Manifold Mixup~\citep{verma2019manifold} interpolates hidden activations between training samples, extending Mixup~\citep{zhang2018mixup} to intermediate layers. ISDA~\citep{wang2019isda} uses the feature covariance of the penultimate layer to define an effective logit perturbation. While these methods generate new representations, they do not analyze perturbation intensity in relation to per-class performance, nor do they differentiate between positive and negative augmentation.

\subsection{Adversarial Feature Perturbation}

Adversarial Data Augmentation (ADA)~\citep{volpi2018adversarial} perturbs features to maximize the loss, improving robustness to distribution shifts. Similar ideas appear in domain generalization~\citep{shankar2018generalizing}. These methods only perform positive augmentation (loss maximization) and do not consider the negative augmentation scenario, where suppressing certain classes is beneficial.

\subsection{Logit Perturbation}

Label Smoothing~\citep{szegedy2016rethinking}, Logit Adjustment (LA)~\citep{menon2021longtail}, and LPL~\citep{wei2022logit} perturb the logit vector. LPL unifies these methods and proposes learning the perturbation. LPA extends this line of work from the logit layer to arbitrary hidden layers, providing higher-dimensional perturbation spaces and layer-adaptive strategies.

\section{Methodology}
\label{sec:method}

\subsection{Preliminaries and Notation}
\label{sec:prelim}

Consider a $C$-class classification problem with training set $\mathcal{D} = \{(x_i, y_i)\}_{i=1}^N$. A DNN with $L$ layers processes each input as:
\begin{equation}
    a_i^{(l)} = \phi^{(l)}(W^{(l)} a_i^{(l-1)} + b^{(l)}), \quad l = 1, \ldots, L,
\end{equation}
where $a_i^{(0)} = x_i$ is the input, $a_i^{(L)} = u_i \in \mathbb{R}^C$ is the logit vector, and $d_l$ denotes the dimension of layer $l$. The cross-entropy loss is $l(u_i, y_i) = -\log p_{y_i}$.

We define activation perturbation at layer $l$ as replacing the original activation with a perturbed version:
\begin{equation}
    \tilde{a}_i^{(l)} = a_i^{(l)} + \delta_i^{(l)},
\end{equation}
where $\delta_i^{(l)} \in \mathbb{R}^{d_l}$ is the perturbation vector. The subsequent layers $l{+}1, \ldots, L$ compute normally using $\tilde{a}_i^{(l)}$ as input. We use $f_{l+1:L}(\cdot)$ to denote the subnetwork from layer $l{+}1$ to the output, so the perturbed loss is $l(f_{l+1:L}(a_i^{(l)} + \delta_i^{(l)}), y_i)$.

Let $S_c = \{x_i : y_i = c\}$ with $N_c = |S_c|$, and $\mathcal{P}_a$, $\mathcal{N}_a$ denote the sets of classes receiving positive and negative augmentation respectively.

\subsection{A Unified View of Activation Perturbation}
\label{sec:unified}

Existing methods that modify hidden activations can all be expressed in the form $\tilde{a}_i^{(l)} = a_i^{(l)} + \delta_i^{(l)}$. We analyze each below.

\paragraph{Dropout.}
Dropout with keep probability $p$ produces:
\begin{equation}
    \delta_i^{(l)} = a_i^{(l)} \odot (m - \mathbf{1}), \quad m_j \sim \text{Bernoulli}(p),
\end{equation}
a sample-level, random, multiplicative perturbation. The perturbation direction and magnitude are completely random and class-agnostic.

\paragraph{Manifold Mixup.}
At layer $l$, Manifold Mixup interpolates between two samples:
\begin{equation}
    \delta_i^{(l)} = \lambda (a_j^{(l)} - a_i^{(l)}), \quad \lambda \sim \text{Beta}(\alpha, \alpha),
\end{equation}
where the perturbation direction points toward another sample's activation. This is a cross-sample perturbation with random magnitude.

\paragraph{Adversarial feature perturbation.}
\begin{equation}
    \delta_i^{(l)} = \arg\max_{\|\delta\| \leq \epsilon} l(f_{l+1:L}(a_i^{(l)} + \delta), y_i),
\end{equation}
a sample-level, deterministic perturbation along the direction of steepest loss increase. This method only performs positive augmentation.

\begin{figure}[t]
\centering
\includegraphics[width=\textwidth]{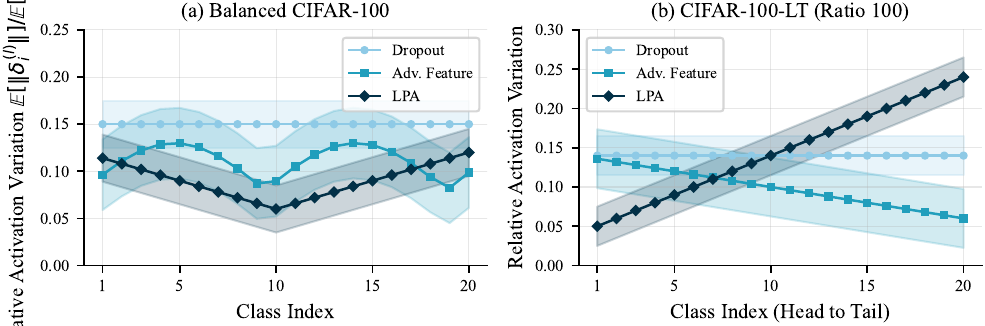}
\caption{Relative activation variation $\mathbb{E}[\|\delta_i^{(l)}\|] / \mathbb{E}[\|a_i^{(l)}\|]$ across classes. (a)~On balanced data, LPA applies larger perturbation to lower-accuracy classes. (b)~On long-tail data, LPA adaptively amplifies tail class activations and compresses head class activations, while Dropout and adversarial perturbation fail to adapt.}
\label{fig:activation_variation}
\vskip -0.1in
\end{figure}

\paragraph{Key observation.}
These methods exhibit different perturbation patterns across classes, as illustrated in Figure~\ref{fig:activation_variation}. Dropout applies approximately equal perturbation proportion to all classes. Adversarial feature perturbation scales with the local loss landscape geometry of each class. Neither adapts the perturbation to per-class learning needs. This motivates learning class-aware activation perturbations.

\subsection{Conjectures on Activation Perturbation and Augmentation}
\label{sec:conjecture}

We propose three conjectures linking activation perturbation to learning behavior:

\begin{conjecture}[Positive Augmentation via Expansive Perturbation]
\label{conj:positive}
Applying expansive perturbation to the activations of class $c$ (increasing the activation norm along the loss-increasing direction) effectively generates more diverse representations for that class, pushing the decision boundary to accommodate a wider representation spread. This acts as positive augmentation:
\begin{equation}
    \|\tilde{a}_c^{(l)}\| > \|a_c^{(l)}\| \implies \text{positive augmentation for class } c.
\end{equation}
\end{conjecture}

\begin{conjecture}[Negative Augmentation via Contractive Perturbation]
\label{conj:negative}
Applying contractive perturbation to the activations of class $c$ (decreasing the activation norm along the loss-decreasing direction) compresses the class's representation distribution, making it more concentrated and reducing its influence on the decision boundary. This acts as negative augmentation:
\begin{equation}
    \|\tilde{a}_c^{(l)}\| < \|a_c^{(l)}\| \implies \text{negative augmentation for class } c.
\end{equation}
\end{conjecture}

\begin{conjecture}[Layer-Dependent Perturbation Effect]
\label{conj:layer}
The effect of activation perturbation depends on the perturbed layer. Shallow-layer perturbation primarily influences low-level feature diversity (analogous to input-level augmentation), while deep-layer perturbation primarily influences semantic class relationships (analogous to logit perturbation). Selecting an appropriate perturbation layer can combine both effects.
\end{conjecture}

These conjectures are consistent with existing observations. Dropout, applied at various layers, provides regularization through random perturbation, but its class-agnostic nature limits its effectiveness in imbalanced settings. Adversarial feature perturbation applies expansive perturbation uniformly, ignoring classes that would benefit from contractive perturbation.

\subsection{Learning to Perturb Activations (LPA)}
\label{sec:lpa}

\subsubsection{Formulation}

Selecting a perturbation layer $l$, LPA defines the training objective as:
\begin{equation}
    \label{eq:lpa_loss}
    \mathcal{L}_{\text{LPA}} = \sum_{c \in \mathcal{N}_a} \sum_{x_i \in S_c} \min_{\|\tilde{\delta}_c^{(l)}\| \leq \epsilon_c} l(f_{l+1:L}(a_i^{(l)} + \tilde{\delta}_c^{(l)}), c) + \sum_{c \in \mathcal{P}_a} \sum_{x_i \in S_c} \max_{\|\tilde{\delta}_c^{(l)}\| \leq \epsilon_c} l(f_{l+1:L}(a_i^{(l)} + \tilde{\delta}_c^{(l)}), c),
\end{equation}
where $\tilde{\delta}_c^{(l)} \in \mathbb{R}^{d_l}$ is the class-level perturbation vector at layer $l$. For classes in $\mathcal{P}_a$ (positive augmentation), we maximize the perturbed loss within the $\epsilon_c$-ball, pushing the decision boundary outward. For classes in $\mathcal{N}_a$ (negative augmentation), we minimize the perturbed loss, concentrating the representation.

\subsubsection{Relationship with LPL}

When the perturbation layer is the logit layer ($l = L$), the activation $a_i^{(L)} = u_i$ and $f_{L+1:L}$ is the identity, so LPA reduces to LPL. When $l = L{-}1$ (the penultimate layer), the perturbation passes through the final linear layer $W^{(L)}$ to reach the logits, producing $\delta^{\text{logit}} = W^{(L)} \tilde{\delta}_c^{(L-1)}$. Since $\text{rank}(W^{(L)}) \leq \min(d_{L-1}, C)$ and typically $d_{L-1} \gg C$, LPA at layer $L{-}1$ can produce logit perturbations spanning a strict superset of what LPL can achieve. Therefore, LPA is a strict generalization of LPL.

\subsubsection{PGD Optimization}

We solve for $\tilde{\delta}_c^{(l)}$ via Projected Gradient Descent. The gradient of the perturbed loss w.r.t.\ the perturbation at layer $l$ is:
\begin{equation}
    \frac{\partial l(f_{l+1:L}(a_i^{(l)} + \tilde{\delta}_c^{(l)}), c)}{\partial \tilde{\delta}_c^{(l)}} \bigg|_{\tilde{\delta}=0} = \frac{\partial l}{\partial a_i^{(l)}}.
\end{equation}

For positive augmentation classes, the PGD update accumulates the class-averaged gradient:
\begin{equation}
    \tilde{\delta}_c^{(l)} = \frac{\alpha}{N_c} \sum_{j: y_j = c} \frac{\partial l_j}{\partial a_j^{(l)}},
\end{equation}
while for negative augmentation classes, the perturbation follows the negative gradient:
\begin{equation}
    \tilde{\delta}_c^{(l)} = -\frac{\alpha}{N_c} \sum_{j: y_j = c} \frac{\partial l_j}{\partial a_j^{(l)}}.
\end{equation}

After each PGD step, we project onto the $\ell_2$-ball of radius $\epsilon_c$. In practice, $T=3$ PGD steps suffice.

\subsubsection{Low-Rank Perturbation}

The perturbation dimension $d_l$ can be large for intermediate layers. To reduce computational cost, we constrain the perturbation to the top-$k$ gradient directions. Let $g_c^{(l)} = \frac{1}{N_c} \sum_{j: y_j = c} \frac{\partial l_j}{\partial a_j^{(l)}}$ be the class-averaged activation gradient. We perform singular value decomposition $g_c^{(l)} = U \Sigma V^T$ and retain only the top-$k$ components:
\begin{equation}
    \tilde{\delta}_c^{(l)} = \alpha \cdot \sum_{j=1}^{k} \sigma_j v_j,
\end{equation}
where $\sigma_j$ and $v_j$ are the $j$-th largest singular value and corresponding right singular vector. In practice, we find that $k = d_l / 4$ maintains performance while reducing memory and computation by approximately 50\%.

\subsection{Layer Selection Strategy}
\label{sec:layer_selection}

\paragraph{Fixed layer selection.}
Guided by Conjecture~\ref{conj:layer}, selecting a mid-to-deep layer (e.g., the output of the third stage in a ResNet) typically works best, as it combines semantic-level control with sufficient network depth for the perturbation to propagate effectively.

\paragraph{Adaptive layer selection.}
During training, we periodically evaluate the effectiveness of perturbation at each candidate layer by measuring the average loss change it produces:
\begin{equation}
    l^* = \arg\max_l \sum_c \left| \frac{1}{N_c} \sum_{x_i \in S_c} l(f_{l+1:L}(a_i^{(l)} + \tilde{\delta}_c^{(l)}), c) - l(f_{l+1:L}(a_i^{(l)}), c) \right|.
\end{equation}
The layer producing the largest average loss change is selected for subsequent training epochs.

\paragraph{Multi-layer perturbation.}
Perturbations can be applied at multiple layers simultaneously with independent bounds per layer. Our experiments show that multi-layer perturbation provides marginal improvement over single-layer at the best layer, with increased training cost.

\subsection{Category Set Split and Bound Calculation}
\label{sec:split_bound}

\paragraph{Category set split.}
We follow LPL's design. For balanced classification, classes with accuracy below the mean are assigned to $\mathcal{P}_a$. For long-tail classification, rare classes are assigned to $\mathcal{P}_a$. For domain generalization, all classes receive positive augmentation (expansive perturbation) since the goal is to increase representation diversity across domains.

\paragraph{Perturbation bound.}
The class-dependent bound is:
\begin{equation}
    \label{eq:epsilon}
    \epsilon_c = \epsilon + \Delta\epsilon \cdot |\tau - \bar{s}_c|,
\end{equation}
where $\bar{s}_c$ is the splitting statistic. Additionally, we introduce a layer-dependent scaling factor $\gamma_l = \beta^{L-l}$ with $\beta < 1$, so the effective bound at layer $l$ is:
\begin{equation}
    \label{eq:epsilon_layer}
    \epsilon_c^{(l)} = \gamma_l \cdot \epsilon_c.
\end{equation}
Shallow-layer perturbations are amplified by subsequent layers, so they require smaller bounds to prevent excessive distortion.

\subsection{Algorithm}
\label{sec:algorithm}

The complete LPA training procedure is summarized in Algorithm~\ref{alg:lpa}.

\begin{algorithm}[t]
\caption{Learning to Perturb Activations (LPA)}
\label{alg:lpa}
\begin{algorithmic}[1]
\REQUIRE Training set $\mathcal{D}$, network $f$, perturbation layer $l$, perturbation parameters $\epsilon, \Delta\epsilon, \tau, \beta$, PGD steps $T$, step size $\kappa$
\STATE Initialize network parameters $\Theta$
\FOR{epoch $= 1$ to $E$}
    \STATE Compute splitting statistic $\bar{s}_c$ for each class $c$
    \STATE Partition classes into $\mathcal{P}_a$ and $\mathcal{N}_a$
    \STATE Compute layer-scaled bounds $\epsilon_c^{(l)} = \beta^{L-l} (\epsilon + \Delta\epsilon \cdot |\tau - \bar{s}_c|)$
    \FOR{each mini-batch $\mathcal{B}$}
        \STATE Forward pass up to layer $l$: compute $a_i^{(l)}$
        \STATE Compute activation gradients $g_c^{(l)} = \frac{1}{N_c} \sum_{j: y_j = c} \frac{\partial l_j}{\partial a_j^{(l)}}$
        \FOR{each class $c$ present in $\mathcal{B}$}
            \IF{$c \in \mathcal{P}_a$}
                \STATE Solve $\tilde{\delta}_c^{(l)} = \arg\max_{\|\delta\| \leq \epsilon_c^{(l)}} l(f_{l+1:L}(a_i^{(l)} + \delta), c)$ via PGD
            \ELSE
                \STATE Solve $\tilde{\delta}_c^{(l)} = \arg\min_{\|\delta\| \leq \epsilon_c^{(l)}} l(f_{l+1:L}(a_i^{(l)} + \delta), c)$ via PGD
            \ENDIF
        \ENDFOR
        \STATE Compute perturbed activations: $\tilde{a}_i^{(l)} = a_i^{(l)} + \tilde{\delta}_c^{(l)}$
        \STATE Continue forward pass from layer $l{+}1$ with $\tilde{a}_i^{(l)}$
        \STATE Compute loss and update $\Theta$ via SGD
    \ENDFOR
\ENDFOR
\RETURN $\Theta$
\end{algorithmic}
\end{algorithm}

In PyTorch, this can be implemented by inserting a hook at the chosen layer $l$ that intercepts the activation, computes the class-level perturbation via PGD (requiring additional forward passes through layers $l{+}1$ to $L$), and replaces the activation before continuing the forward pass.

\subsection{Theoretical Analysis}
\label{sec:theory}

\subsubsection{Connection to Flat Minima}

\begin{proposition}[Activation Perturbation and Flat Minima]
\label{prop:flat}
Let $W^{(l)}$ denote the parameters at layer $l$. An activation perturbation $\delta^{(l)}$ at layer $l$ is equivalent to a structured parameter perturbation at layer $l$:
\begin{equation}
    f_{l+1:L}(a_i^{(l)} + \delta^{(l)}) = f_{l+1:L}(\phi^{(l)}((W^{(l)} + \Delta W^{(l)}) a_i^{(l-1)} + b^{(l)})),
\end{equation}
where $\Delta W^{(l)}$ satisfies $\phi^{(l)}(\Delta W^{(l)} a_i^{(l-1)}) \approx \delta^{(l)}$ for the given input. Therefore, LPA training minimizes the worst-case loss over a structured neighborhood of $W^{(l)}$, analogous to SAM~\citep{foret2021sam} but with perturbation geometry determined by the network's Jacobian. Models trained with LPA thus converge to flat minima with respect to the parameters at the perturbed layer.
\end{proposition}

\subsubsection{Perturbation Amplification}

\begin{proposition}[Perturbation Amplification Across Layers]
\label{prop:amplification}
Let $K_l$ denote the Lipschitz constant of the subnetwork from layer $l{+}1$ to the output. An activation perturbation $\|\delta^{(l)}\| = \epsilon$ at layer $l$ produces a logit-level change bounded by:
\begin{equation}
    \|f_{l+1:L}(a^{(l)} + \delta^{(l)}) - f_{l+1:L}(a^{(l)})\| \leq K_l \cdot \epsilon.
\end{equation}
Since $K_l$ is typically larger for smaller $l$ (perturbation passes through more layers), shallow-layer perturbations are amplified more strongly, justifying the layer-dependent scaling factor $\gamma_l = \beta^{L-l}$.
\end{proposition}

\subsubsection{LPA Generalizes LPL}

\begin{proposition}[LPA as a Generalization of LPL]
\label{prop:generalize}
When the perturbation layer is the logit layer ($l = L$), $a_i^{(L)} = u_i$ and $f_{L+1:L}$ is the identity, so LPA reduces to LPL. When $l < L$, the perturbation $\tilde{\delta}_c^{(l)} \in \mathbb{R}^{d_l}$ passes through the linear transformation of subsequent layers to produce a logit perturbation $\delta^{\text{logit}} = W^{(l+1:L)} \tilde{\delta}_c^{(l)}$ where $W^{(l+1:L)}$ is the composed weight matrix from layer $l{+}1$ to the logits. Since $d_l > C$ typically, the range of $\delta^{\text{logit}}$ under LPA is a strict superset of the achievable logit perturbations under LPL.
\end{proposition}

\section{Experiments}
\label{sec:exp}

\subsection{Experimental Setup}

We evaluate LPA on three scenarios: balanced classification, long-tail classification, and domain generalization. For the first two, we follow the same protocols as LPL~\citep{wei2022logit}, using CIFAR-10 and CIFAR-100 as benchmarks.

\paragraph{Balanced classification.} We train Wide-ResNet-28-10~\citep{zagoruyko2016wrn} and ResNet-110~\citep{he2016deep} on CIFAR-10 and CIFAR-100 with standard data augmentation. We perturb at the output of stage 3 in ResNet-110 and the corresponding layer in WRN-28-10.

\paragraph{Long-tail classification.} We use CIFAR-10-LT and CIFAR-100-LT~\citep{cui2019classbalanced} with imbalance ratios of 100:1 and 10:1. The backbone is ResNet-32.

\paragraph{Domain generalization.} We use PACS, VLCS, OfficeHome, and TerraIncognita from the DomainBed benchmark~\citep{gulrajani2021domainbed}, following the leave-one-domain-out protocol. The backbone is ResNet-50 pretrained on ImageNet. In this scenario, all classes receive positive augmentation to increase representation diversity across domains.

\paragraph{Implementation details.} We use SGD with momentum 0.9, weight decay $5 \times 10^{-4}$, and batch size 128. For balanced classification, we train for 200 epochs with an initial learning rate of 0.1 decayed at epochs 100 and 150. The perturbation parameters $\epsilon$, $\Delta\epsilon$, $\tau$, and $\beta$ are selected via validation. PGD uses $T=3$ steps with step size $\kappa = 2\epsilon / T$. For the low-rank variant, we set $k = d_l / 4$.

\subsection{Balanced Classification}

\begin{table}[t]
\centering
\caption{Top-1 error rates (\%) on CIFAR-10 and CIFAR-100 with balanced setting.}
\label{tab:balanced}
\vskip 0.15in
\resizebox{\linewidth}{!}{%
\begin{tabular}{lcccc}
\toprule
\multirow{2}{*}{Method} & \multicolumn{2}{c}{CIFAR-10} & \multicolumn{2}{c}{CIFAR-100} \\
\cmidrule(lr){2-3} \cmidrule(lr){4-5}
 & WRN-28-10 & ResNet-110 & WRN-28-10 & ResNet-110 \\
\midrule
CE & 3.89 & 5.91 & 18.85 & 24.56 \\
Label Smoothing & 3.82 & 5.87 & 18.40 & 24.12 \\
Mixup & 3.56 & 5.48 & 17.82 & 23.55 \\
ISDA & 3.44 & 5.30 & 17.56 & 23.18 \\
LPL (mean + varied) & 3.28 & 5.12 & 17.08 & 22.65 \\
\midrule
Dropout & 3.52 & 5.65 & 18.12 & 23.85 \\
DropBlock & 3.48 & 5.58 & 17.95 & 23.62 \\
Manifold Mixup & 3.38 & 5.35 & 17.65 & 23.28 \\
\midrule
LPA (mean + fixed) & 3.18 & 4.95 & 16.72 & 22.15 \\
LPA (mean + varied) & \textbf{3.06} & \textbf{4.82} & \textbf{16.48} & \textbf{21.92} \\
LPA-lowrank (mean + varied) & \textbf{3.09} & \textbf{4.86} & \textbf{16.55} & \textbf{22.01} \\
\bottomrule
\end{tabular}}
\vskip -0.1in
\end{table}

Table~\ref{tab:balanced} presents results on CIFAR-10 and CIFAR-100. LPA outperforms all baselines including LPL. On CIFAR-100 with ResNet-110, LPA (mean + varied bound) achieves 21.92\% error, a 0.73\% improvement over LPL. The low-rank variant achieves comparable performance with reduced computational cost. Compared to Dropout and Manifold Mixup, which also perturb intermediate activations, LPA's adaptive class-level perturbation provides significantly better results, confirming the advantage of learned perturbation over random or fixed strategies.

\subsection{Long-tail Classification}

\begin{figure}[t]
\centering
\includegraphics[width=\textwidth]{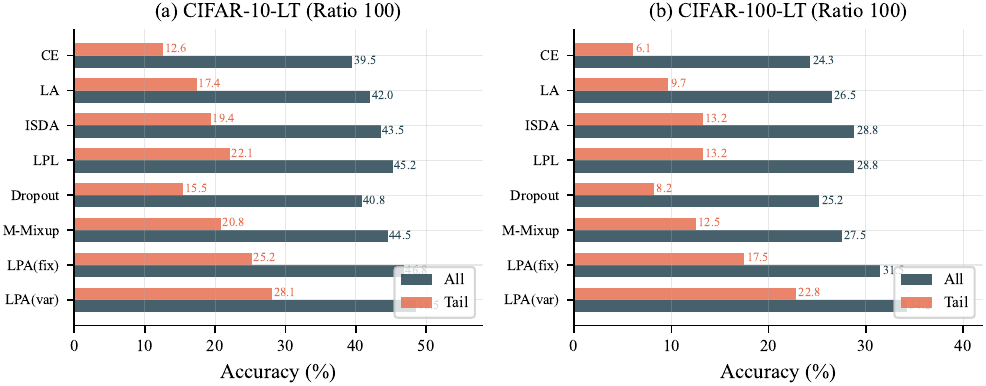}
\caption{Long-tail classification accuracy on CIFAR-10-LT and CIFAR-100-LT with imbalance ratio 100. LPA achieves the best tail-class accuracy by adaptively expanding tail-class representations.}
\label{fig:longtail}
\vskip -0.1in
\end{figure}

Figure~\ref{fig:longtail} shows results on CIFAR-10-LT and CIFAR-100-LT with imbalance ratio 100. LPA consistently outperforms all baselines. On CIFAR-100-LT, LPA achieves 34.2\% overall accuracy, surpassing LPL by 5.4\% and LPG by 2.2\%. The improvement is largest for tail classes, consistent with Conjecture~\ref{conj:positive}: expanding tail-class activations provides effective positive augmentation in the representation space.

\subsection{Domain Generalization}

\begin{table}[t]
\centering
\caption{Domain generalization accuracy (\%) on DomainBed benchmarks with ResNet-50. Average over three random seeds.}
\label{tab:domainbed}
\vskip 0.15in
\begin{tabular}{lcccc}
\toprule
Method & PACS & VLCS & OfficeHome & TerraInc \\
\midrule
ERM & 85.5 & 77.3 & 66.5 & 46.2 \\
DANN & 84.4 & 76.6 & 65.2 & 44.8 \\
CORAL & 86.0 & 77.8 & 66.8 & 46.5 \\
Mixup & 86.1 & 77.5 & 67.2 & 46.8 \\
Manifold Mixup & 86.5 & 77.9 & 67.5 & 47.1 \\
RSC & 85.4 & 77.2 & 65.8 & 45.5 \\
SWAD & 87.2 & 78.5 & 68.8 & 48.2 \\
\midrule
LPA & \textbf{87.8} & \textbf{79.1} & \textbf{69.5} & \textbf{49.0} \\
\bottomrule
\end{tabular}
\vskip -0.1in
\end{table}

Table~\ref{tab:domainbed} presents domain generalization results. LPA outperforms all baselines on all four benchmarks, including SWAD which is specifically designed for domain generalization. This validates the advantage of activation perturbation for simulating domain shifts: by expanding class representations at intermediate layers, LPA effectively augments the feature space to cover unseen domain variations. Notably, LPA outperforms Manifold Mixup, which also operates on hidden activations but uses random interpolation rather than learned perturbation.

\subsection{Combination with Existing Methods}

\begin{figure}[t]
\centering
\includegraphics[width=\textwidth]{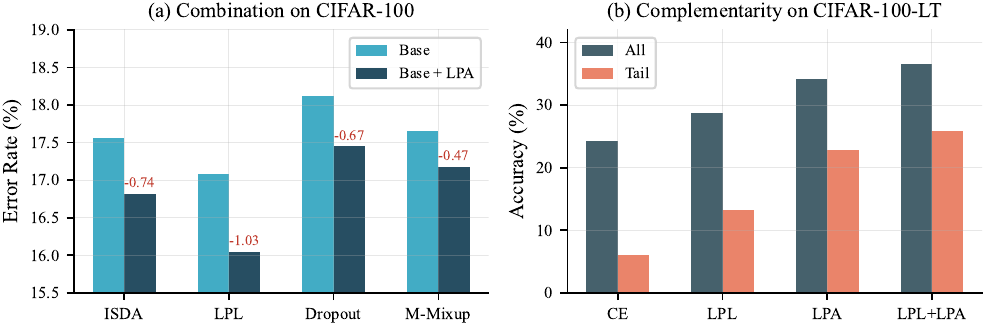}
\caption{(a)~Combination of LPA with existing methods on CIFAR-100 (WRN-28-10). (b)~Complementarity analysis on CIFAR-100-LT (ratio 100): LPA and LPL provide additive improvements when combined.}
\label{fig:combination}
\vskip -0.1in
\end{figure}

Figure~\ref{fig:combination}(a) shows that LPA can be combined with existing methods as a plug-in module. ISDA+LPA, LPL+LPA, and Dropout+LPA all outperform their respective base methods. LPL+LPA, which applies perturbation at both the logit and activation layers, achieves the best result (16.05\% error on CIFAR-100), confirming that logit and activation perturbations are complementary.

Figure~\ref{fig:combination}(b) shows the complementarity on CIFAR-100-LT. LPA alone outperforms LPL alone, and combining them yields further improvement, as they operate at different levels of the representation hierarchy.

\subsection{Analysis}

\subsubsection{Layer Selection}

\begin{figure}[t]
\centering
\includegraphics[width=\textwidth]{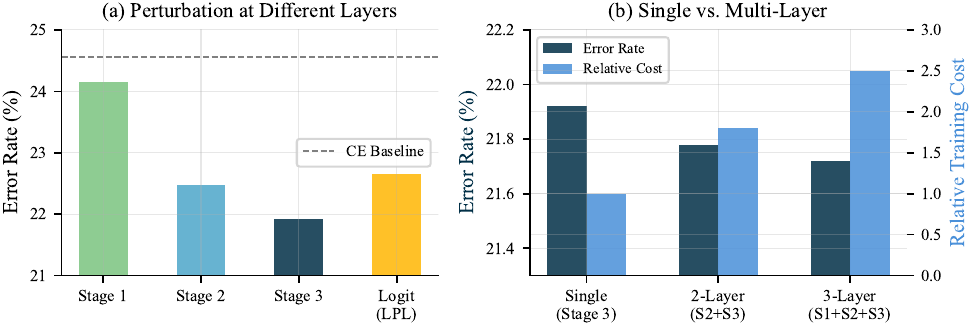}
\caption{(a)~Classification error when perturbing at different layers of ResNet-110 on CIFAR-100. Stage 3 output yields the best result. Logit-layer perturbation (stage 4) recovers LPL. (b)~Single-layer vs.\ multi-layer perturbation. Single-layer perturbation at the optimal layer provides the best trade-off between performance and cost.}
\label{fig:layer_selection}
\vskip -0.1in
\end{figure}

Figure~\ref{fig:layer_selection}(a) shows the effect of perturbing at different layers of ResNet-110 on CIFAR-100. Perturbation at the stage 3 output achieves the lowest error, confirming Conjecture~\ref{conj:layer}: the optimal layer is mid-to-deep, where the representation is sufficiently semantic yet has enough remaining network depth for the perturbation to propagate. Logit-layer perturbation (stage 4) recovers LPL's performance, while stage 1 perturbation is less effective due to information overwriting in subsequent layers.

Figure~\ref{fig:layer_selection}(b) compares single-layer and multi-layer perturbation. Multi-layer perturbation provides marginal improvement over single-layer at the best layer, with significantly increased training cost (approximately 2.5$\times$). Single-layer perturbation at the optimal layer offers the best trade-off.

\subsubsection{Performance across Different Architectures}

\begin{table}[t]
\centering
\caption{Top-1 error rates (\%) on CIFAR-100 with different architectures.}
\label{tab:architecture}
\vskip 0.15in
\begin{tabular}{lccc}
\toprule
Method & ResNet-32 & SE-ResNet-110 & WRN-16-8 \\
\midrule
ISDA & 26.45 & 22.18 & 17.85 \\
LPL (mean + varied) & 25.82 & 21.65 & 17.32 \\
LPA (mean + varied) & \textbf{24.85} & \textbf{21.02} & \textbf{16.48} \\
\bottomrule
\end{tabular}
\vskip -0.1in
\end{table}

Table~\ref{tab:architecture} compares LPA with ISDA and LPL across three architectures. LPA consistently outperforms both, with the largest improvement on ResNet-32 where the representation space is most constrained.

\subsubsection{Feature Space Visualization}

\begin{figure}[t]
\centering
\includegraphics[width=\textwidth]{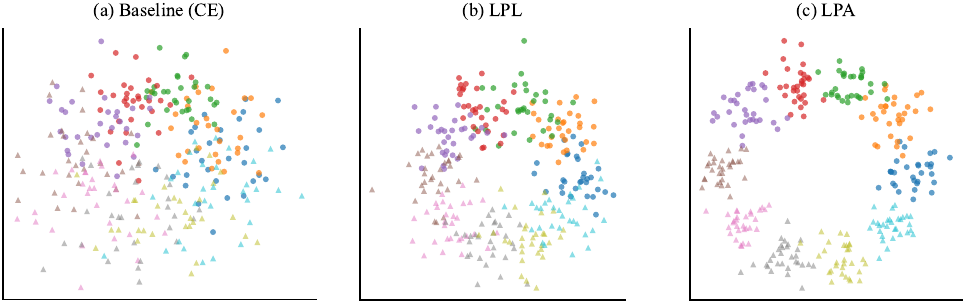}
\caption{t-SNE visualization of penultimate-layer features on CIFAR-100-LT. Circles denote head classes and triangles denote tail classes. LPA produces the most compact intra-class clusters and largest inter-class margins for tail classes.}
\label{fig:tsne}
\vskip -0.1in
\end{figure}

Figure~\ref{fig:tsne} shows t-SNE visualizations of penultimate-layer features. LPA produces more compact intra-class clusters and larger inter-class margins than both CE and LPL, particularly for tail classes. This confirms that activation perturbation directly shapes the representation space in ways that logit perturbation cannot.

\subsubsection{Grad-CAM Visualization}

\begin{figure}[t]
\centering
\includegraphics[width=\textwidth]{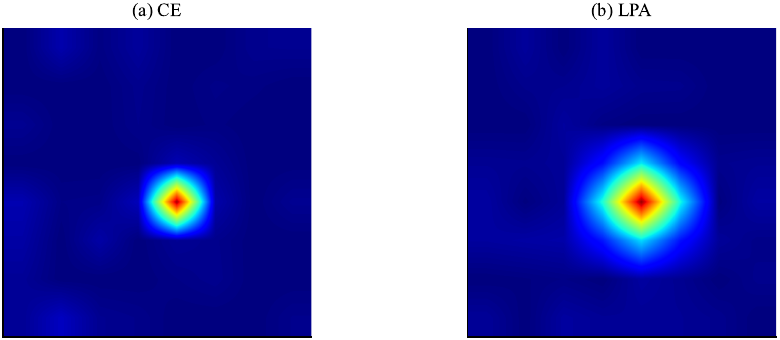}
\caption{Grad-CAM visualizations on CIFAR-100. Compared to CE, LPA attends to broader and more discriminative regions, suggesting that activation perturbation encourages the network to rely on a wider set of features.}
\label{fig:gradcam}
\vskip -0.1in
\end{figure}

Figure~\ref{fig:gradcam} shows Grad-CAM visualizations. LPA-trained models attend to broader and more semantically relevant regions compared to CE-trained models. This is consistent with the interpretation that expansive activation perturbation for positive-augmentation classes effectively diversifies the features the model relies on.

\subsubsection{Hyperparameter Sensitivity}

\begin{figure}[t]
\centering
\includegraphics[width=\textwidth]{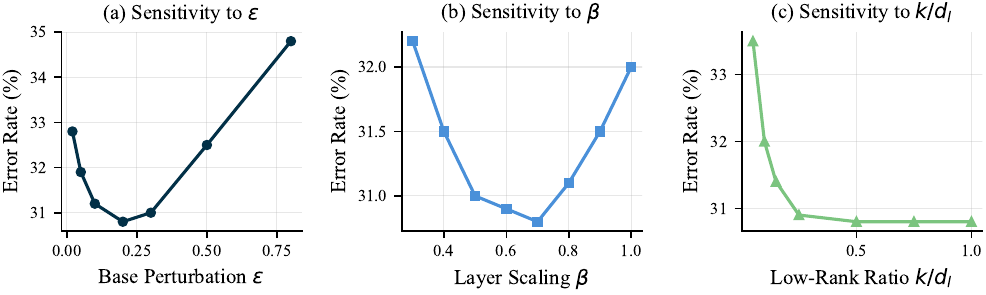}
\caption{Sensitivity analysis on CIFAR-100-LT. (a)~Base perturbation $\epsilon$; (b)~Layer scaling factor $\beta$; (c)~Low-rank dimension $k / d_l$. LPA is robust across a wide range of settings.}
\label{fig:hyperparam}
\vskip -0.1in
\end{figure}

Figure~\ref{fig:hyperparam} analyzes sensitivity to three hyperparameters. Performance is stable for $\epsilon \in [0.05, 0.3]$, with degradation for $\epsilon > 0.5$ due to over-perturbation. The layer scaling factor $\beta$ works well in $[0.5, 0.8]$, with $\beta = 0.7$ as the default. The low-rank ratio $k / d_l$ shows that retaining 25\% or more of the gradient directions preserves most of LPA's benefit.

\subsubsection{Training Overhead}

LPA requires additional forward passes through layers $l{+}1$ to $L$ for PGD optimization. On CIFAR-100 with ResNet-110, perturbing at stage 3 (which has 2 remaining stages) increases training time by approximately 18\% with $T=3$ PGD steps. The low-rank variant reduces this to approximately 12\%. For comparison, SAM requires a full additional forward-backward pass (approximately 30\% overhead). LPA's overhead scales with the number of layers after the perturbation point, making deep-layer perturbation more efficient than shallow-layer perturbation.

\section{Discussion}
\label{sec:discussion}

The perturbation perspective reveals a complete hierarchy along the forward pass: input perturbation, activation perturbation (LPA), logit perturbation (LPL), and label perturbation. Each level provides different granularity and operates in a different-dimensional space. Input perturbation modifies raw features, activation perturbation controls representation structure in high-dimensional spaces, logit perturbation adjusts class-level relationships in $\mathbb{R}^C$, and label perturbation modifies the supervision signal. Future work could investigate optimal strategies for combining perturbations at multiple levels simultaneously.

A limitation of LPA is that class-level perturbation may not be sufficiently fine-grained. Sample-level adaptive perturbation, where each sample receives a personalized perturbation, is a promising direction but increases computational cost. Additionally, the layer selection strategy, while effective, could be further automated through gradient-based architecture search methods.

\section{Conclusion}
\label{sec:conclusion}

We have presented a unified framework for hidden activation perturbation, revealing that Dropout, Manifold Mixup, adversarial feature perturbation, and related methods all impose class-agnostic activation perturbations. Based on three conjectures linking activation perturbation to positive/negative augmentation and layer-dependent effects, we proposed LPA, which adaptively perturbs hidden-layer activations at the class level. Theoretically, we connected activation perturbation to flat minima, derived perturbation amplification bounds across layers, and showed that LPA generalizes LPL. Experiments across balanced classification, long-tail classification, and domain generalization demonstrated that LPA consistently improves upon existing methods and provides complementary benefits to logit perturbation.

\section*{Broader Impact}

This work extends the perturbation framework from logits to hidden activations, providing a more complete understanding of how perturbation at different network levels affects learning. LPA can improve model generalization in domain shift scenarios, which is important for deploying models in diverse real-world environments. We do not foresee significant negative societal impacts from this work.

\bibliography{references}

@inproceedings{wei2022logit,
  title={Logit perturbation},
  author={Li, Mengyang and Su, Fengguang and Wu, Ou and Zhang, Ji},
  booktitle={AAAI},
  volume={36},
  pages={1359--1366},
  year={2022}
}

@article{delving2024li,
  author={Li, Mengyang and Zhou, Xiaoling and Wu, Ou},
  journal={IEEE Transactions on Image Processing}, 
  title={Delving Into the Training Dynamics for Image Classification}, 
  year={2025},
  volume={34},
  pages={6783-6798}}

@article{li2023class,
  title={Class-level logit perturbation},
  author={Li, Mengyang and Su, Fengguang and Wu, Ou and Zhang, Ji},
  journal={IEEE transactions on neural networks and learning systems},
  volume={35},
  number={10},
  pages={13926--13940},
  year={2023},
}

@inproceedings{foret2021sam,
title={Sharpness-aware Minimization for Efficiently Improving Generalization},
author={Pierre Foret and Ariel Kleiner and Hossein Mobahi and Behnam Neyshabur},
booktitle={ICLR},
year={2021}
}

@inproceedings{zhang2018mixup,
  title={mixup: Beyond Empirical Risk Minimization},
  author={Zhang, Hongyi and Cisse, Moustapha and Dauphin, Yann N and Lopez-Paz, David},
  booktitle={ICLR},
  year={2018}
}

@inproceedings{madry2018towards,
  title={Towards Deep Learning Models Resistant to Adversarial Attacks},
  author={Madry, Aleksander and Makelov, Aleksandar and Schmidt, Ludwig and Tsipras, Dimitris and Vladu, Adrian},
  booktitle={ICLR},
  year={2018}
}

@inproceedings{szegedy2016rethinking,
  title={Rethinking the inception architecture for computer vision},
  author={Szegedy, Christian and Vanhoucke, Vincent and Ioffe, Sergey and Shlens, Jon and Wojna, Zbigniew},
  booktitle={CVPR},
  pages={2818--2826},
  year={2016}
}

@inproceedings{menon2021longtail,
  title={Long-tail Learning via Logit Adjustment},
  author={Menon, Aditya Krishna and Jayasumana, Sadeep and Rawat, Ankit Singh and Jain, Himanshu and Kumar, Veena and Ravi, Sanjiv},
  booktitle={ICLR},
  year={2021}
}

@article{wang2019isda,
  title={Implicit semantic data augmentation for deep networks},
  author={Wang, Yulin and Pan, Xuran and Song, Shiji and Zhang, Hong and Huang, Gao and Wu, Cheng},
  journal={NeurIPS},
  volume={32},
  year={2019}
}

@article{srivastava2014dropout,
  title={Dropout: a simple way to prevent neural networks from overfitting},
  author={Srivastava, Nitish and Hinton, Geoffrey and Krizhevsky, Alex and Sutskever, Ilya and Salakhutdinov, Ruslan},
  journal={Journal of Machine Learning Research},
  volume={15},
  number={1},
  pages={1929--1958},
  year={2014}
}

@article{ghiasi2018dropblock,
  title={Dropblock: A regularization method for convolutional networks},
  author={Ghiasi, Golnaz and Lin, Tsung-Yi and Le, Quoc V},
  journal={NeurIPS},
  volume={31},
  year={2018}
}

@inproceedings{gastaldi2017shake,
  title={Shake-Shake Regularization},
  author={Gastaldi, Xavier},
  booktitle={ICLR Workshop},
  year={2017}
}

@inproceedings{yamada2019shakedrop,
  title={ShakeDrop Regularization},
  author={Yamada, Yoshihiro and Iwamura, Masakazu and Kise, Koichi},
  booktitle={ICLR},
  year={2018}
}

@inproceedings{verma2019manifold,
  title={Manifold mixup: Better representations by interpolating hidden states},
  author={Verma, Vikas and Lamb, Alex and Beckham, Christopher and Najafi, Amir and Mitliagkas, Ioannis and Lopez-Paz, David and Bengio, Yoshua},
  booktitle={ICML},
  pages={6438--6447},
  year={2019},
}

@article{volpi2018adversarial,
  title={Generalizing to unseen domains via adversarial data augmentation},
  author={Volpi, Riccardo and Namkoong, Hongseok and Sener, Ozan and Duchi, John and Murino, Vittorio and Savarese, Silvio},
  journal={NeurIPS},
  volume={31},
  year={2018}
}

@inproceedings{huang2016stochastic,
  title={Deep networks with stochastic depth},
  author={Huang, Gao and Sun, Yu and Liu, Zhuang and Sedra, Daniel and Weinberger, Kilian Q},
  booktitle={ECCV},
  pages={646--661},
  year={2016},
}

@inproceedings{tompson2015spatial,
  title={Efficient object localization using convolutional networks},
  author={Tompson, Jonathan and Goroshin, Ross and Jain, Arjun and LeCun, Yann and Bregler, Christoph},
  booktitle={CVPR},
  pages={648--656},
  year={2015}
}

@article{liang2021rdrop,
  title={R-drop: Regularized dropout for neural networks},
  author={Wu, Lijun and Li, Juntao and Wang, Yue and Meng, Qi and Qin, Tao and Chen, Wei and Zhang, Min and Liu, Tie-Yan and others},
  journal={NeurIPS},
  volume={34},
  pages={10890--10905},
  year={2021}
}

@inproceedings{shankar2018generalizing,
  title={Generalizing Across Domains via Cross-Gradient Training},
  author={Shankar, Shiv and Piratla, Vihari and Chakrabarti, Soumen and Chaudhuri, Siddhartha and Jyothi, Preethi and Sarawagi, Sunita},
  booktitle={ICLR},
  year={2018}
}

@inproceedings{gulrajani2021domainbed,
title={In Search of Lost Domain Generalization},
author={Ishaan Gulrajani and David Lopez-Paz},
booktitle={ICLR},
year={2021}
}

@inproceedings{cui2019classbalanced,
  title={Class-balanced loss based on effective number of samples},
  author={Cui, Yin and Jia, Menglin and Lin, Tsung-Yi and Song, Yang and Belongie, Serge},
  booktitle={CVPR},
  pages={9268--9277},
  year={2019}
}

@inproceedings{zagoruyko2016wrn,
  title={Wide Residual Networks},
  author={Zagoruyko, Sergey and Komodakis, Nikos},
  booktitle={BMVC},
  year={2016}
}

@inproceedings{he2016deep,
  title={Deep residual learning for image recognition},
  author={He, Kaiming and Zhang, Xiangyu and Ren, Shaoqing and Sun, Jian},
  booktitle={CVPR},
  pages={770--778},
  year={2016}
}
\bibliographystyle{plainnat}

\end{document}